\documentclass{article}
\usepackage{ijcai17}
\usepackage{times}
\usepackage{helvet}
\usepackage{courier}
\usepackage{setspace}

\usepackage{color}
\usepackage{comment}
\usepackage{url}

\usepackage{booktabs}
\usepackage{amsfonts}
\usepackage{amsmath}
\usepackage{amssymb}
\usepackage{makecell}
\usepackage{subcaption}
\usepackage{multirow}
\usepackage{pbox}

\usepackage[flushleft]{threeparttable}

\usepackage{graphicx,xspace}
\usepackage{epstopdf}
\usepackage{stfloats}
\def\st#1{~}
\def\inv{\vspace{-0.1cm}}

\begin{document}
%
\title{Multilingual Knowledge Graph Embeddings for\\ Cross-lingual Knowledge Alignment}

\author{Muhao Chen$^1$, Yingtao Tian$^2$, Mohan Yang$^1$, Carlo Zaniolo$^1$\\
\{muhaochen, yang, zaniolo\}@cs.ucla.edu; yittian@cs.stonybrook.edu\\
Department of Computer Science, UCLA$^1$\\
Department of Computer Science, Stony Brook University$^2$\\
}
\maketitle

\begin{abstract}
Many recent works have demonstrated the benefits of knowledge graph embeddings in completing monolingual knowledge graphs.
Inasmuch as related knowledge bases are built in several different languages, achieving cross-lingual knowledge alignment will help people in constructing a \mbox{coherent} knowledge base, and assist machines in dealing with different expressions of entity relationships across diverse human languages.
Unfortunately, achieving this highly desirable cross-lingual alignment
by human labor is very costly and error-prone.
Thus,
we propose MTransE, a translation-based model for multilingual knowledge graph embeddings, to provide a simple and automated solution.
By encoding
entities and relations of each language in a separated embedding space, MTransE
provides transitions for each
embedding vector to its cross-lingual counterparts in other spaces, while preserving the functionalities of monolingual embeddings.
We deploy three different techniques to represent cross-lingual transitions, namely axis calibration, translation vectors, and linear transformations, and derive five
variants for MTransE using
different loss functions.
Our models can be
trained on partially aligned graphs, where just a small portion of triples are aligned with their cross-lingual counterparts.
The experiments on cross-lingual entity matching and triple-wise alignment verification show promising results, with some variants consistently outperforming others on different tasks.
We also explore how MTransE preserves the key properties of its monolingual counterpart TransE.
\end{abstract}

\section{Introduction}

Multilingual knowledge bases such as 
Wikipedia~\cite{url:wiki}, WordNet~\cite{bond2013linking}, and ConceptNet~\cite{speer2013conceptnet} are becoming essential sources of knowledge 
for people and AI-related applications.
These 
knowledge bases are modeled as knowledge graphs that store two aspects of knowledge: the {\em monolingual knowledge} that includes 
entities and 
relations recorded in the form of triples, and the {\em cross-lingual knowledge} that matches the monolingual knowledge among various human languages. \par

The coverage issue of monolingual knowledge has been widely addressed
, and parsing-based techniques for completing monolingual knowledge bases have been well studied in the past~\cite{culotta2004dependency,guodong2005exploring,sun2011semi}. 
More recently, 
much attention has been paid to embedding-based techniques, which 
provide simple methods to encode entities in low-dimensional embedding spaces and capture relations as means of translations among entity vectors.
Given a triple $(h, r, t)$ where $r$ is the relation between entities $h$ and $t$, then $h$ and $t$ are represented as two $k$-dimensional vectors $\mathbf{h}$ and $\mathbf{t}$, respectively. A function $f_{r}(\mathbf{h},\mathbf{t})$ is used to measure the plausibility of $(h, r, t)$, which also implies the transformation $\mathbf{r}$ that characterizes $r$.
Exemplarily, the translation-based model TransE~\cite{bordes2013translating} 
uses the loss function $f_{r}(\mathbf{h},\mathbf{t})=\left \| \mathbf{h}+\mathbf{r}-\mathbf{t} \right \|$~\footnote{Hereafter, $\|\cdot\|$ means $l_1$ or $l_2$ norm unless explicitly specified.}, where $\mathbf{r}$ is characterized as a translation vector learnt from the latent connectivity patterns in the knowledge graph. This model provides 
a flexible way of predicting
a missing item in a triple, or verifying the validity of a generated triple. 
Other works like TransH~\cite{wang2014knowledge} and TransR~\cite{lin2015learning}, introduce different loss functions that represent the relational translation in other forms, and have\st{ already }achieved promising results in 
completing the knowledge graphs. 
\par 

While embedding-based techniques 
can help improve the completeness of
monolingual knowledge, the problem 
of applying these techniques on 
cross-lingual knowledge remains largely unexplored.
Such knowledge, including {\em inter-lingual links} (ILLs) that match the same entities, and {\em triple-wise alignment} (TWA) that represents the same relations, is
very helpful in synchronizing different language-specific \mbox{versions} of a knowledge base that evolve independently, as needed to 
further improve applications built on knowledge bases, such as Q\&A systems, semantic Web, and Web search.
In spite of its importance, this cross-lingual knowledge \mbox{remains} largely intact.
In fact, in 
the most successful knowledge base Wikipedia, 
we find that ILLs cover less than 15\% entity alignment.
\par

Leveraging knowledge graph embeddings to cross-lingual knowledge no doubt provides a generic way 
to help \mbox{extract} and apply such knowledge. 
However, it is a non-trivial task to find a tractable technique to capture the cross-lingual transitions\footnote{We use the word \emph{transition} here to differentiate from the relational translations among entities in translation-based methods.}. Such transitions are more difficult to capture than relational translations for several 
reasons: 
(i) a cross-lingual transition has a far larger domain 
than any monolingual relational translation;
(ii) it applies on both entities and relations, which have incoherent vocabularies among different languages;
(iii) the known alignment for training such transitions usually accounts for a small percentage of a knowledge base.
Moreover, 
the characterization of monolingual knowledge graph structures has to be well-preserved to ensure the 
correct representation of the knowledge to be aligned.

To address 
the above issues, 
we propose
a multilingual knowledge graph embedding model {\em MTransE}, that
\mbox{learns} the multilingual knowledge graph structure using 
a combination of two component models, namely {\em knowledge model} and {\em alignment model}.
The knowledge model encodes 
entities and relations in a language-specific 
version of knowledge graph.
We explore 
the method that
organizes each language-specific version in a \mbox{separated} embedding space, 
in which
MTransE adopts 
TransE as the knowledge model. 
On top of that, the alignment model learns cross-lingual transitions for both entities and relations across different embedding spaces, where the following three representations of 
cross-lingual alignment are considered: distance-based axis calibration, translation vectors, and linear transformations. 
Thus, we obtain five variants of MTransE based on different loss functions, and identify the best variant by comparing them on cross-lingual \mbox{alignment} tasks using two partially aligned trilingual graphs constructed from Wikipedia triples. We also show that MTransE performs as well as its monolingual counterpart TransE on monolingual tasks.

The rest of the paper is organized as follows. We first discuss the related work, and then introduce our approach in the section that follows. After that we present the experimental results
, and conclude the paper in the last section.

\newcommand{\stitle}[1]{\vspace{0.3ex}\noindent{\bf #1}}

\inv
\section{Related Work} \label{sec:related}


While, at the best of our knowledge, there is no previous work 
on learning multilingual knowledge graph embeddings, we will describe next three lines of work which are closely related to this topic.
\par

\stitle{Knowledge Graph Embeddings.} Recently, significant advancement has been made in using the translation-based method to train monolingual knowledge graph embeddings. 
To characterize a triple $(h, r, t)$, models of this family follow a common assumption 
$\mathbf{h}_r + \mathbf{r} \approx \mathbf{t}_r$,
where $\mathbf{h}_r$ and $\mathbf{t}_r$ are either the original vectors of $h$ and $t$, or the transformed vectors under a 
certain transformation w.r.t. relation $r$.
The forerunner
TransE~\cite{bordes2013translating} sets $\mathbf{h}_r$ and $\mathbf{t}_r$ as the original $\mathbf{h}$ and $\mathbf{t}$,
and 
achieves 
promising results in handling 1-to-1 relations. 
Later works improve TransE on multi-mapping relations by introducing
relation-specific transformations on entities
to obtain different $\mathbf{h}_r$ and $\mathbf{t}_r$, including 
projections on 
relation-specific hyperplanes in TransH~\cite{wang2014knowledge},
linear transformations to heterogeneous relation spaces in TransR~\cite{lin2015learning}, dynamic matrices in TransD~\cite{ji2015knowledge}, and other forms
~\cite{jia2016locally,nguyenstranse}.
All these variants of TransE specialize
entity embeddings for different relations,
therefore improving 
knowledge graph completion on multi-mapping relations at the cost of 
increased 
model complexity. Meanwhile translation-based models 
cooperate well with other models. For example, variants of TransE are combined 
with word embeddings to help relation extraction from text~\cite{westonconnecting,zhong2015aligning}. \par

In addition to these
, there are
non-translation-based methods. 
Some of those including UM~\cite{bordes2011learning}, SE \cite{bordes2012joint}, Bilinear \cite{jenatton2012latent}, and HolE \cite{nickel2016holographic}, do not explicitly represent relation embeddings.
Others including neural-based models SLM~\cite{collobert2008unified} and NTN~\cite{socher2013reasoning}, and random-walk-based model {TADW}~\cite{yang2015network}, are expressive and adaptable for both structured and text corpora, but are too complex to be incorporated into an architecture supporting
multilingual knowledge.
\par

\stitle{Multilingual Word Embeddings.}
Several approaches learn multilingual word embeddings
on parallel text corpora. Some of those can be extended to multilingual knowledge graphs, such as LM \cite{mikolov2013exploiting} and CCA \cite{faruqui2014improving} which induce offline transitions among pre-trained monolingual embeddings in forms of linear transformations and canonical component analysis respectively.
These approaches do not adjust the inconsistent vector spaces via calibration or jointly training with the \mbox{alignment} model, thus fail to perform well on knowledge graphs as
the parallelism exists only in small portions.
A better approach OT~\cite{xing2015normalized} jointly learns regularized embeddings and orthogonal transformations, which is however found to be overcomplicated due to the inconsistency of monolingual vector spaces and the large diversity of relations among entities.


\stitle{Knowledge Bases Alignment.} Some projects produce cross-lingual alignment in knowledge bases
at the cost of extensive human involvement and designing
hand-crafted features dedicated to specific applications.
Wikidata \cite{vrandevcic2012wikidata} and
DBpedia \cite{lehmann2015dbpedia} rely on crowdsourcing to create ILLs and relation alignment. YAGO \cite{mahdisoltani2014yago3} mines association rules on known matches, which combines many confident scores and requires extensively fine tuning.
Many other works require sources that are external to the graphs, from well-established schemata or ontologies \cite{nguyen2011multi,suchanek2011paris,rinser2013cross} to entity descriptions \cite{yang2015entity}, which being unavailable to many knowledge bases such as YAGO, WordNet, and ConceptNet \cite{speer2013conceptnet}.
Such approaches also involve complicated model dependencies that are not tractable and reusable.
By contrast, embedding-based methods
are simple and general, require little human involvement,
and generate task-independent features that can contribute to other NLP tasks.
\par

\def\kb{\mathit{KB}}
\def\language{\mathcal{L}}
\def\bhline{\specialrule{.2em}{0em}{0em}}
\newcommand{\bigO}[1]{{\rm O} (#1)\xspace}

\inv
\section{Multilingual Knowledge Graph Embeddings}

We hereby begin our modeling with the formalization of multilingual knowledge graphs.

\inv
\subsection{Multilingual Knowledge Graphs}

In a knowledge base $\kb$, we use $\language$ to denote the set of languages, and $\language^2$ to denote the 2-combination of $\language$ (i.e., the set of \emph{unordered} language pairs). 
For a language $L \in \language$, $G_{L}$ denotes the language-specific knowledge graph of $L$, and $E_{L}$ and $R_{L}$ respectively denote the corresponding vocabularies of entity expression and relation expression.
$T = (h,r,t)$ denotes a triple in $G_{L}$ such that $h, t \in E_{L}$ and $r \in R_{L}$. Boldfaced $\mathbf{h}$, $\mathbf{r}$, $\mathbf{t}$ respectively represent the embedding vectors of head $h$, relation $r$, and tail $t$. For a language pair 
$(L_1, L_2) \in \language^2$, 
$\delta(L_1, L_2)$ denotes the alignment set which contains
the pairs of triples that have already been aligned
between $L_1$ and $L_2$.
For example, across the languages English and French, we may have $\bigl(($State of California, capital city, Sacramento$), ($\'{E}tat de Californie, capitale, Sacramento$)\bigr) \in \delta(\mbox{English, French})$.
The alignment set commonly exists in a small portion in a multilingual knowledge base~\cite{vrandevcic2012wikidata,mahdisoltani2014yago3,lehmann2015dbpedia}, and is one part of knowledge we want to extend.
\par


Our model consists of 
two components that 
learn on the two facets of $\kb$: the knowledge model that encodes the entities and relations from each language-specific graph structure, and the alignment model that learns the cross-lingual transitions from the existing 
alignment. 
We define a model for each language pair from $\language^2$ that has a non-empty alignment set. Thus, for a $\kb$ with more than two languages, a set of models composes the solution. In the following, we use a language pair ($L_i$, $L_j$) $\in \language^2$ as an example to describe how we define each component of a model.
\par


\inv
\subsection{Knowledge Model}
For each language $L \in \language$, a dedicated $k$-dimensional embedding space $\mathbb{R}_{L}^{k}$ is assigned for vectors of $ E_{L}$ and $ R_{L}$, where $\mathbb{R}$ is the field of real numbers. 
We adopt the basic translation-based method of TransE
for each involved language, which benefits the cross-lingual tasks by representing embeddings uniformly in different contexts of relations.
Therefore its loss function is given as below:
\begin{equation*}
S_K = \sum_{L \in \{L_i, L_j\}} \sum_{(h,r,t) \in G_L} \left \| \mathbf{h} + \mathbf{r} - \mathbf{t} \right \|
\end{equation*}
It measures the plausibility of all given triples. By minimizing the loss function, the knowledge model preserves monolingual relations among entities, while also acts as a regularizer for the alignment model. Meanwhile, 
the knowledge model partitions the knowledge base into 
disjoint subsets that can be trained in parallel.


\inv
\subsection{Alignment Model}

The objective of the alignment model is to construct the transitions between the vector spaces of $L_i$ and $L_j$. 
Its loss function is given as below:
\begin{equation*}
S_{A} = \sum_{(T, T') \in \delta(L_{i}, L_{j})} S_{a}(T,T')
\end{equation*}
Thereof, the alignment score $S_{a}(T, T')$ iterates through all pairs of aligned triples. Three different techniques to score the alignment are considered: distance-based axis calibration, translation vectors, and linear transformations. Each of them is based on a different assumption, and constitutes different forms of $S_{a}$ alongside.

\stitle{Distance-based Axis Calibration.}
This type of \mbox{alignment} models 
penalize the alignment based on the distances of cross-lingual counterparts. Either of the following two scorings can be adopted to the model.
\begin{equation*}
S_{a_{1}} = \left \| \mathbf{h} - \mathbf{h}' \right \| + \left \| \mathbf{t} - \mathbf{t}' \right \|
\end{equation*}
$S_{a_{1}}$ regulates that correctly aligned multilingual \mbox{expressions} of the same entity tend to have close embedding vectors. Thus by minimizing the loss function that involves $S_{a_{1}}$ on known pairs of aligned triples, the alignment model adjusts axes of embedding spaces towards the goal of coinciding the vectors of the same entity in different languages.
\begin{equation*}
S_{a_{2}} = \left \| \mathbf{h} - \mathbf{h}' \right \| + \left \| \mathbf{r} - \mathbf{r}' \right \| + \left \| \mathbf{t} - \mathbf{t}' \right \|
\end{equation*}
$S_{a_{2}}$ overlays the penalty of relation alignment to $S_{a_{1}}$ to explicitly converge coordinates of the same relation. \par

The alignment models based on axis calibration assume analogous spatial emergence of items in each language. Therefore, it realizes the cross-lingual transition by carrying forward the vector of a given entity or relation from the space of the original language to that of the other language.

\stitle{Translation Vectors.}
This model encodes cross-lingual transitions into vectors. It 
consolidates alignment into graph structures and characterizes cross-lingual transitions as regular relational translations. Hence $S_{a_{3}}$ as below is derived. 
\begin{equation*}
S_{a_{3}} = \left \| \mathbf{h} + \mathbf{v}_{ij}^{e} - \mathbf{h}' \right \| + \left \| \mathbf{r} + \mathbf{v}_{ij}^{r} - \mathbf{r}' \right \| + \left \| \mathbf{t} + \mathbf{v}_{ij}^{e} - \mathbf{t}' \right \|
\end{equation*}
Thereof $\mathbf{v}_{ij}^{e}$ and $\mathbf{v}_{ij}^{r}$ are respectively deployed as the entity-dedicated and relation-dedicated translation vectors between $L_i$ and $L_j$, such that we have $\mathbf{e} + \mathbf{v}_{ij}^{e} \approx \mathbf{e'}$ for embedding vectors $\mathbf{e}$, $\mathbf{e}'$ of the same entity $e$ expressed in both 
languages, and $\mathbf{r} + \mathbf{v}_{ij}^{r} \approx \mathbf{r'}$ for those of the same relation. We deploy two translation vectors instead of one, because there are far more distinct entities than relations, and using 
one vector easily leads to imbalanced signals from relations. \par

Such a model obtains a cross-lingual transition of 
an embedding vector by adding the corresponding translation vector.
Moreover
, it is easy to see that $\mathbf{v}_{ij}^{e} = -\mathbf{v}_{ji}^{e}$ and $\mathbf{v}_{ij}^{r} = - \mathbf{v}_{ji}^{r}$ hold. 
Therefore, as we obtain the translation vectors from $L_i$ to $L_j$, we can always use the same vectors to translate in the opposite direction.

\stitle{Linear Transformations.}
The last category of alignment models deduce linear transformations between embedding spaces. $S_{a_4}$ as below learns a $k\times k$ square matrix $\mathbf{M}_{ij}^{e}$ as a linear transformation on entity vectors from $L_i$ to $L_j$, given $k$ as the dimensionality of the embedding spaces.
\begin{equation*}
S_{a_{4}} = \left \| \mathbf{M}_{ij}^{e} \mathbf{h} - \mathbf{h}' \right \| + \left \| \mathbf{M}_{ij}^{e} \mathbf{t} - \mathbf{t}' \right \|
\end{equation*}
$S_{a_5}$ additionally brings in a second linear transformation $\mathbf{M}_{ij}^{r}$ for relation vectors, which is of the same shape as $\mathbf{M}_{ij}^{e}$. The use of a different matrix is again due to different redundancy of entities and relations.
\begin{equation*}
S_{a_{5}} = \left \| \mathbf{M}_{ij}^{e} \mathbf{h} - \mathbf{h}' \right \| + \left \| \mathbf{M}_{ij}^{r} \mathbf{r} - \mathbf{r}' \right \| + \left \| \mathbf{M}_{ij}^{e} \mathbf{t} - \mathbf{t}' \right \|
\end{equation*}
Unlike 
axis calibration, linear-transformation-based alignment model treats cross-lingual transitions as the topological transformation of embedding spaces without assuming the similarity of spatial emergence. \par

The cross-lingual transition of a vector is obtained by applying the corresponding linear transformation. It is noteworthy that, regularization of embedding vectors in the training process (which will be introduced soon after) ensures the invertibility
of the linear transformations such that ${\mathbf{M}_{ij}^{e}}^{-1} = {\mathbf{M}_{ji}^{e}}$ and ${\mathbf{M}_{ij}^{r}}^{-1} = {\mathbf{M}_{ji}^{r}}$. Thus the transition in the revert direction is always enabled even though the model only learns the transformations of one direction.

\inv
\subsection{Variants of MTransE}

Combining the above two component models, MTransE minimizes the following loss function 
$
J = S_K + \alpha S_A
$,
where $\alpha$ is a hyperparameter that weights $S_K$ and $S_A$.

As we have given out five variants of the alignment model, each of which correspondingly defines its specific way of computing cross-lingual transitions of embedding vectors.
We denote Var$_k$ as the variant of MTransE that adopts the $k$-th alignment model which employs $S_{a_k}$.
In practice, the searching of a cross-lingual counterpart for a source is always done by querying the nearest neighbor from the result point of the cross-lingual transition.
We denote function $\tau_{ij}$ that maps a cross-lingual transition of a vector from $L_i$ to $L_j$, or simply $\tau$ in a bilingual context. 
As stated,
the solution in a multi-lingual scenario consists of a set of models of the same variant defined on every language pair in $\language^2$.
Table~\ref{tbl:complexity} summarizes the model complexity, the definition of cross-lingual transitions, and the complexity of searching a cross-lingual counterpart for each variant.

\begin{table}[t]
\centering
\caption{
Summary of model variants.
}
\label{tbl:complexity}
\begin{threeparttable}
\vspace{-1em}
{
\scriptsize
\begin{tabular} {c|c@{\hspace{1em}}c@{\hspace{1em}}c}
\bhline
Var&Model Complexity&Cross-lingual Transition&Search Complexity\\
\bhline
Var$_1$&$\bigO{n_ekl+n_rkl}$&\makecell{$\tau_{ij}(\mathbf{e})=\mathbf{e}$\\$\tau_{ij}(\mathbf{r})=\mathbf{r}$}&\makecell{$\bigO{n_ek}$\\ $\bigO{n_rk}$}\\
\hline
Var$_2$&$\bigO{n_ekl+n_rkl}$&\makecell{$\tau_{ij}(\mathbf{e})=\mathbf{e}$\\$\tau_{ij}(\mathbf{r})=\mathbf{r}$}&\makecell{$\bigO{n_ek}$\\ $\bigO{n_rk}$}\\
\hline
Var$_3$&\makecell{${\rm O}(n_ekl+n_rkl$\\$~~~~~~~~~~~~~~~~+kl^{2})$}
&\makecell{$\tau_{ij}(\mathbf{e})=\mathbf{e}+\mathbf{v}_{ij}^{e}$\\$\tau_{ij}(\mathbf{r})=\mathbf{r}+\mathbf{v}_{ij}^{r}$}&\makecell{$\bigO{n_ek}$\\ $\bigO{n_rk}$}\\
\hline
Var$_4$&\makecell{${\rm O}(n_ekl+n_rkl$\\$~~~~~~~~+0.5k^{2}l^{2})$}
&\makecell{$\tau_{ij}(\mathbf{e})=\mathbf{M}_{ij}^{e} \mathbf{e}$\\$\tau_{ij}(\mathbf{r})=\mathbf{M}_{ij}^{e} \mathbf{r}$}&\makecell{$\bigO{n_ek^2+n_ek}$\\$\bigO{n_rk^2+n_rk}$}\\
\hline
Var$_5$&\makecell{${\rm O}(n_ekl+n_rkl$\\$~~~~~~~~~~~~~+k^{2}l^{2})$}
&\makecell{$\tau_{ij}(\mathbf{e})=\mathbf{M}_{ij}^{e} \mathbf{e}$\\$\tau_{ij}(\mathbf{r})=\mathbf{M}_{ij}^{r} \mathbf{r}$}&\makecell{$\bigO{n_ek^2+n_ek}$\\$\bigO{n_rk^2+n_rk}$}\\
\bhline
\end{tabular}}
\footnotesize
\begin{tablenotes}
\scriptsize
\item Notation: $\mathbf{e}$ and $\mathbf{r}$ are respectively the vectors of an entity $e$ and a relation $r$, $k$ is the dimension 
    of the embedding spaces, $l$ is the cardinality of $\language$, $n_e$ and $n_r$ are respectively the number of entities and the number of relations, where $n_e \gg n_r$.
\end{tablenotes}
\end{threeparttable}
\vspace{-1.5em}
\end{table}

\inv\inv
\subsection{Training}

We optimize the loss function 
using on-line stochastic gradient descent~\cite{wilson2003general}.
At each step, we update the parameter $\theta$ by setting
$
\theta \leftarrow  \theta - \lambda \nabla_\theta J
$,
where $\lambda$ is the learning rate.
Instead of directly updating $J$, our implementation optimizes $S_K$ and $\alpha S_A$ alternately.
In detail, at each epoch we optimize $\theta \leftarrow  \theta - \lambda \nabla_\theta S_K$ and $\theta \leftarrow  \theta - \lambda \nabla_\theta \alpha S_A$ in separated groups of steps.



We enforce the constraint that the $l_2$ norm of any entity embedding vector is 1,
thus regularize embedding vectors to a unit spherical surface.
This constraint is employed in the literature
\cite{bordes2013translating,bordes2014open,jenatton2012latent} and has two important effects:
(i) it helps avoid the case where the training process trivially minimizes the loss function by shrinking the norm of embedding vectors, and
(ii) it implies the invertibility of the linear transformations \cite{xing2015normalized} for Var$_4$ and Var$_5$.

We initialize vectors by drawing from a uniform distribution on the unit spherical surface,
and initialize matrices using random orthogonal initialization \cite{saxe2013exact}.
Negative sampling is not employed in training, which we find does not noticeably affect the \mbox{results}.
\def\hits{\mathit{Hits}\mbox{@}10}
\def\mean{\mathit{Mean}}

\inv
\section{Experiments}

In this section, we evaluate the proposed methods on two cross-lingual tasks: cross-lingual entity matching, and triple-wise alignment verification.
We also conduct experiments on two monolingual tasks. Besides, a \emph{case study} with knowledge alignment examples is included in the Appendix of \cite{chen2016arxiv}.

\begin{table}[t]
\centering
\caption{Statistics of the WK3l data sets.}
\label{tbl:statistics}
\vspace{-1em}
\scriptsize
\begin{tabular}{c|cccc}
\bhline
Data set&\#En triples&\#Fr triples&\#De triples&\#Aligned triples\\
\bhline
WK3l-15k&203,502&170,605&145,616&\makecell{En-Fr:16,470\\En-De:37,170}\\
\hline
WK3l-120k&1,376,011&767,750&391,108&\makecell{En-Fr:124,433\\En-De:69,413}\\
\bhline
\end{tabular}
\vspace{-1em}
\end{table}

\begin{table}[t]
\centering
\caption{Number of entity inter-lingual links (ILLs).}
\label{tbl:ills}
\vspace{-1em}
\scriptsize
\begin{tabular}{c|cccc}
\bhline
Data Set&En-Fr&Fr-En&En-De&De-En\\
\bhline
WK3l-15k&3,733&3,815&1,840&1,610\\
\hline
WK3l-120k&42,413&41,513&7,567&5,921\\
\bhline
\end{tabular}
\vspace{-1.5em}
\end{table}

\begin{table*}[t]
\centering
\caption{Cross-lingual entity matching result.
}
\label{tbl:matching}
\vspace{-1em}
\scriptsize
\begin{tabular}{c*{12}{|c}}
\bhline
Data Set&\multicolumn{8}{c|}{WK3l-15k}&\multicolumn{4}{c}{WK3l-120k}\\
\hline
Aligned Languages&\multicolumn{2}{c|}{En-Fr}&\multicolumn{2}{c|}{Fr-En}&\multicolumn{2}{c|}{En-De}&\multicolumn{2}{c|}{De-En}&\multicolumn{1}{c|}{En-Fr}&\multicolumn{1}{c|}{Fr-En}&\multicolumn{1}{c|}{En-De}&\multicolumn{1}{c}{De-En}\\
\hline
Metric&\multicolumn{1}{c}{Hits@10}&\multicolumn{1}{c|}{Mean}&\multicolumn{1}{c}{Hits@10}&\multicolumn{1}{c|}{Mean}&\multicolumn{1}{c}{Hits@10}&\multicolumn{1}{c|}{Mean}&\multicolumn{1}{c}{Hits@10}&\multicolumn{1}{c|}{Mean}&\multicolumn{1}{c|}{Hits@10}&\multicolumn{1}{c|}{Hits@10}&\multicolumn{1}{c|}{Hits@10}&\multicolumn{1}{c}{Hits@10}\\
\bhline
LM&\multicolumn{1}{c}{12.31}&\multicolumn{1}{c|}{3621.17}&\multicolumn{1}{c}{10.42}&\multicolumn{1}{c|}{3660.98}&\multicolumn{1}{c}{22.17}&\multicolumn{1}{c|}{5891.13}&\multicolumn{1}{c}{15.21}&\multicolumn{1}{c|}{6114.08}&\multicolumn{1}{c|}{11.74}&\multicolumn{1}{c|}{14.26}&\multicolumn{1}{c|}{24.52}&\multicolumn{1}{c}{13.58}\\
CCA&\multicolumn{1}{c}{20.78}&\multicolumn{1}{c|}{3094.25}&\multicolumn{1}{c}{19.44}&\multicolumn{1}{c|}{3017.90}&\multicolumn{1}{c}{26.46}&\multicolumn{1}{c|}{5550.89}&\multicolumn{1}{c}{22.30}&\multicolumn{1}{c|}{5855.61}&\multicolumn{1}{c|}{19.47}&\multicolumn{1}{c|}{12.85}&\multicolumn{1}{c|}{25.54}&\multicolumn{1}{c}{20.39}\\
OT&\multicolumn{1}{c}{44.97}&\multicolumn{1}{c|}{508.39}&\multicolumn{1}{c}{40.92}&\multicolumn{1}{c|}{461.18}&\multicolumn{1}{c}{44.47}&\multicolumn{1}{c|}{155.47}&\multicolumn{1}{c}{49.24}&\multicolumn{1}{c|}{145.47}&\multicolumn{1}{c|}{38.91}&\multicolumn{1}{c|}{37.19}&\multicolumn{1}{c|}{38.85}&\multicolumn{1}{c}{34.21}\\
\hline
Var${_1}$&\multicolumn{1}{c}{51.05}&\multicolumn{1}{c|}{470.29}&\multicolumn{1}{c}{46.64}&\multicolumn{1}{c|}{436.47}&\multicolumn{1}{c}{48.67}&\multicolumn{1}{c|}{146.13}&\multicolumn{1}{c}{50.60}&\multicolumn{1}{c|}{167.02}&\multicolumn{1}{c|}{38.58}&\multicolumn{1}{c|}{36.52}&\multicolumn{1}{c|}{42.06}&\multicolumn{1}{c}{47.79}\\
Var${_2}$&\multicolumn{1}{c}{45.25}&\multicolumn{1}{c|}{570.72}&\multicolumn{1}{c}{41.74}&\multicolumn{1}{c|}{565.38}&\multicolumn{1}{c}{46.27}&\multicolumn{1}{c|}{168.33}&\multicolumn{1}{c}{49.00}&\multicolumn{1}{c|}{211.94}&\multicolumn{1}{c|}{31.88}&\multicolumn{1}{c|}{30.84}&\multicolumn{1}{c|}{41.22}&\multicolumn{1}{c}{40.39}\\
Var${_3}$&\multicolumn{1}{c}{38.64}&\multicolumn{1}{c|}{587.46}&\multicolumn{1}{c}{36.44}&\multicolumn{1}{c|}{464.64}&\multicolumn{1}{c}{50.82}&\multicolumn{1}{c|}{125.15}&\multicolumn{1}{c}{52.16}&\multicolumn{1}{c|}{151.84}&\multicolumn{1}{c|}{38.26}&\multicolumn{1}{c|}{36.45}&\multicolumn{1}{c|}{50.48}&\multicolumn{1}{c}{52.24}\\
Var${_4}$&\multicolumn{1}{c}{59.24}&\multicolumn{1}{c|}{\textbf{190.26}}&\multicolumn{1}{c}{\textbf{57.48}}&\multicolumn{1}{c|}{\textbf{199.64}}&\multicolumn{1}{c}{\textbf{66.25}}&\multicolumn{1}{c|}{\textbf{74.62}}&\multicolumn{1}{c}{\textbf{68.53}}&\multicolumn{1}{c|}{\textbf{42.31}}&\multicolumn{1}{c|}{\textbf{48.66}}&\multicolumn{1}{c|}{47.43}&\multicolumn{1}{c|}{57.56}&\multicolumn{1}{c}{63.49}\\
Var${_5}$&\multicolumn{1}{c}{\textbf{59.52}}&\multicolumn{1}{c|}{191.36}&\multicolumn{1}{c}{57.07}&\multicolumn{1}{c|}{204.45}&\multicolumn{1}{c}{60.25}&\multicolumn{1}{c|}{99.48}&\multicolumn{1}{c}{66.03}&\multicolumn{1}{c|}{54.69}&\multicolumn{1}{c|}{45.65}&\multicolumn{1}{c|}{\textbf{47.48}}&\multicolumn{1}{c|}{\textbf{64.22}}&\multicolumn{1}{c}{\textbf{67.85}}\\
\bhline
\end{tabular}
\vspace{-1em}
\end{table*}

\begin{figure*}[t]
\centering
\begin{subfigure}[c]{0.235\textwidth}
\centering
\includegraphics[width=\textwidth]{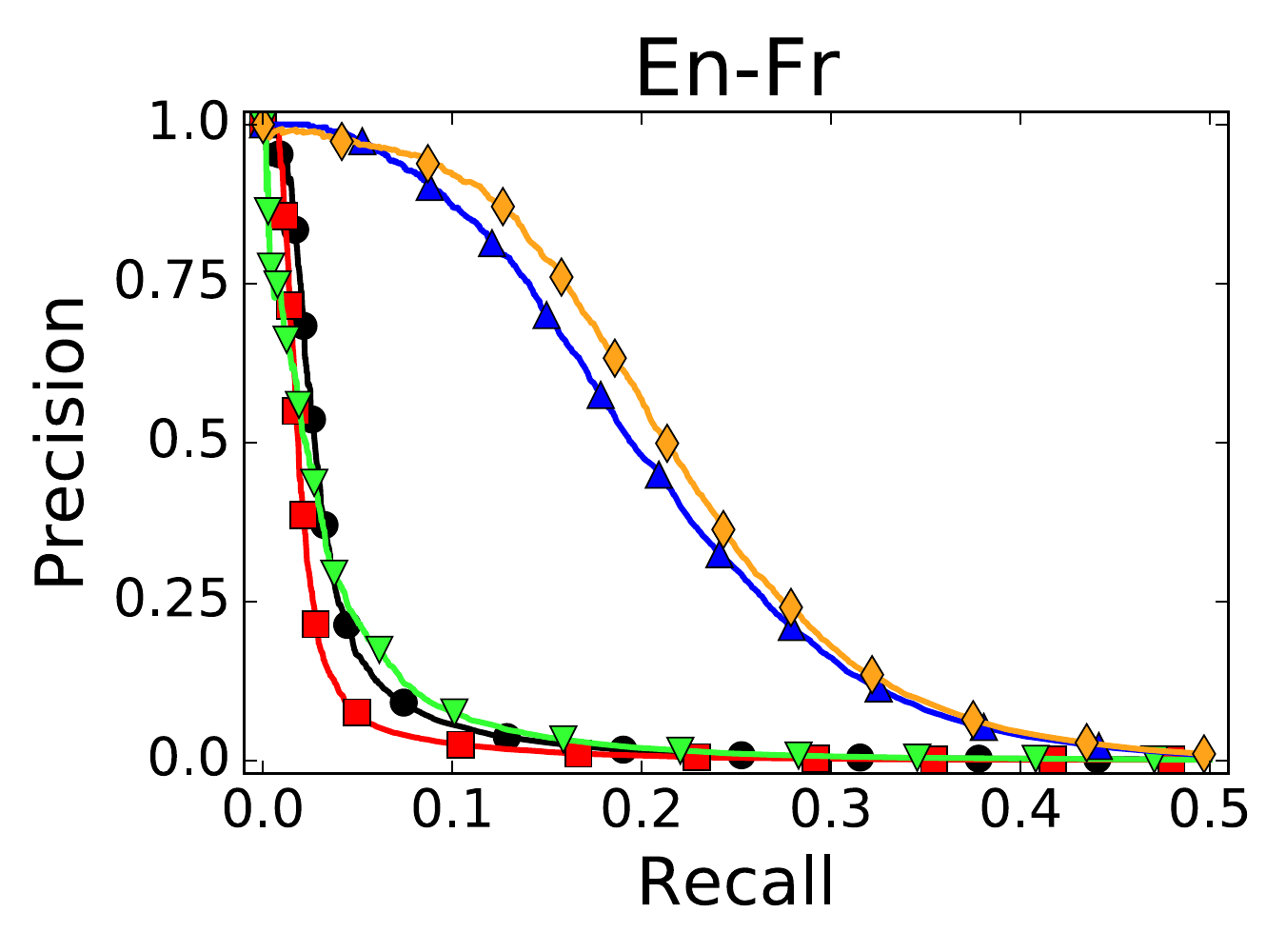}
\end{subfigure}
\hspace{-0.55em}
\begin{subfigure}[c]{0.235\textwidth}
\centering
\includegraphics[width=\textwidth]{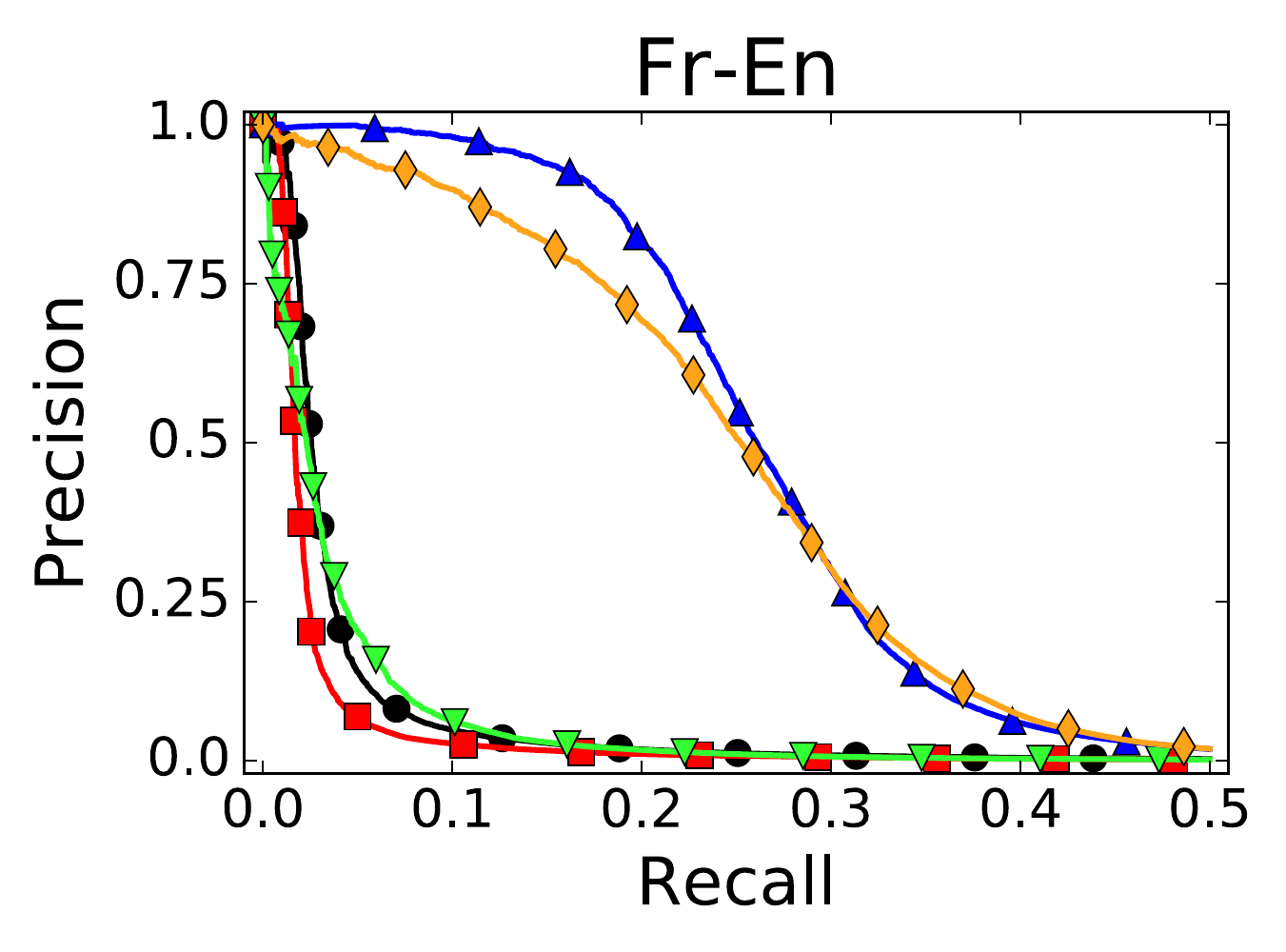}
\end{subfigure}
\hspace{-0.55em}
\begin{subfigure}[c]{0.235\textwidth}
\centering
\includegraphics[width=\textwidth]{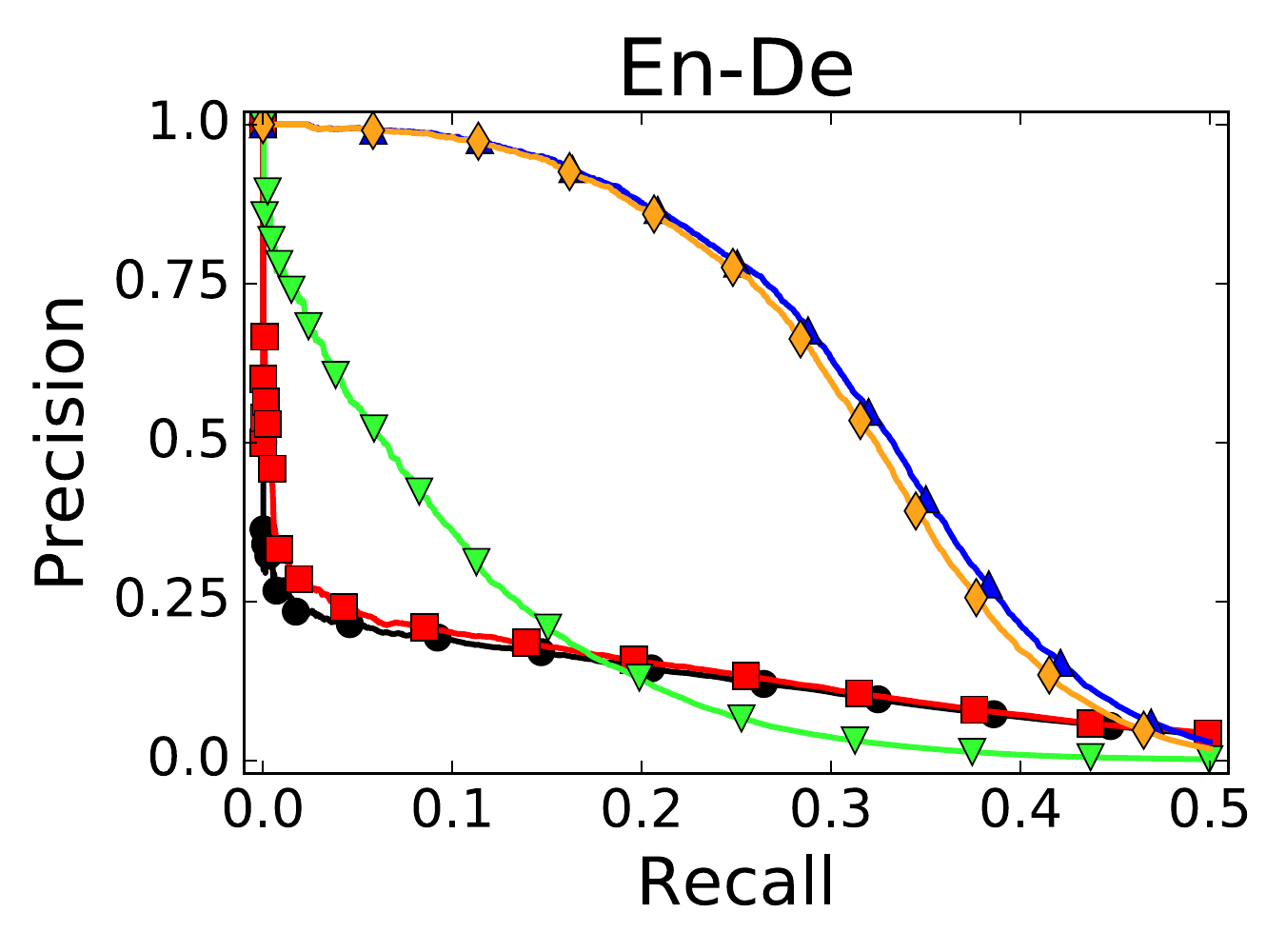}
\end{subfigure}
\hspace{-0.55em}
\begin{subfigure}[c]{0.235\textwidth}
\centering
\includegraphics[width=\textwidth]{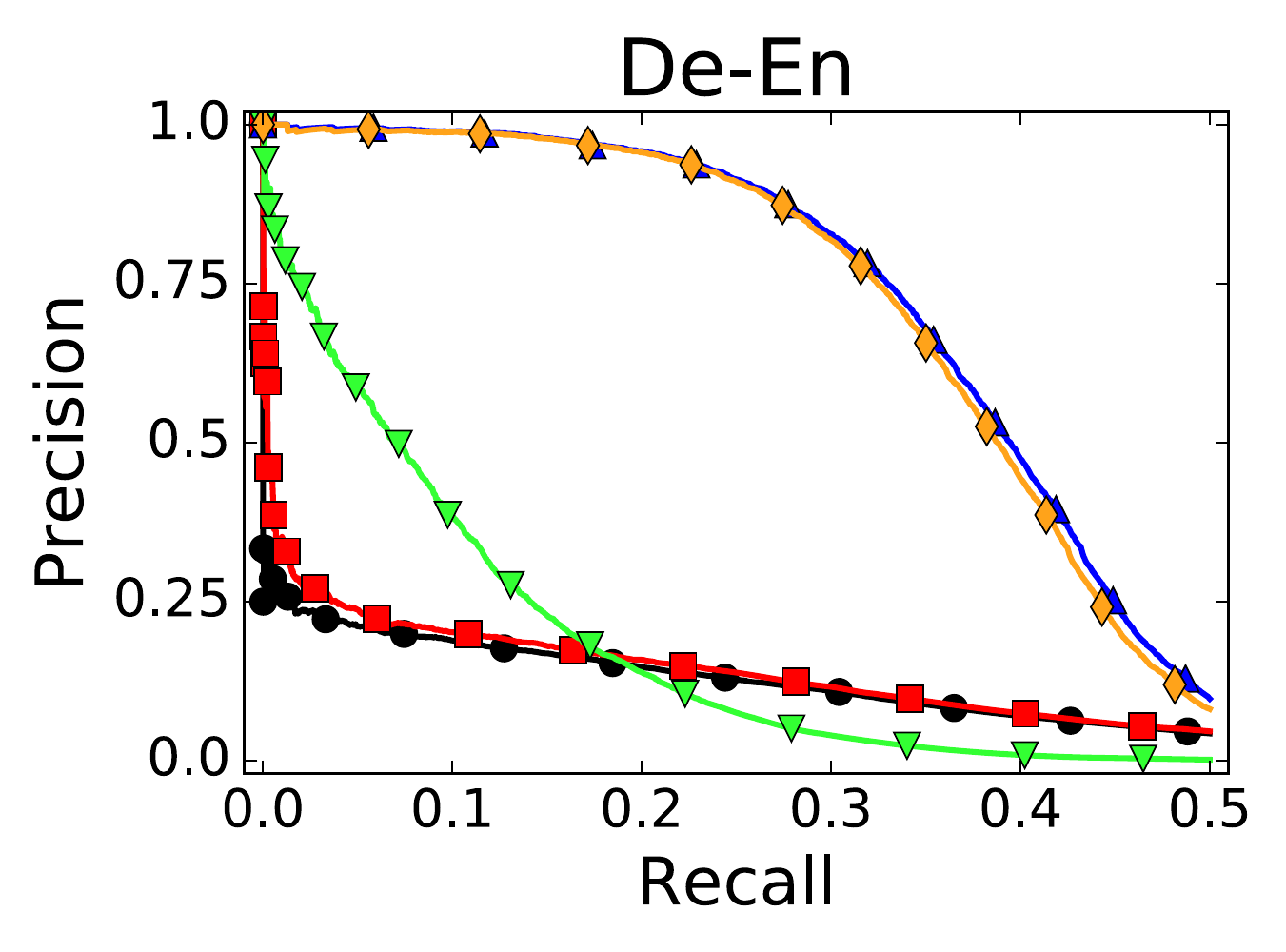}
\end{subfigure}
\hspace{-0.55em}
\begin{subfigure}[c]{0.06\textwidth}
\centering
\includegraphics[width=\textwidth]{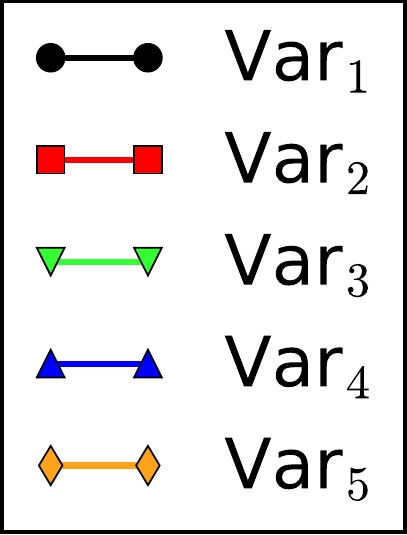}
\end{subfigure}
\vspace{-1em}
\caption{Precision-recall curves for cross-lingual entity matching on WK3l-15k.}
\label{fig:matching}
\vspace{-1.5em}
\end{figure*}


\stitle{Data Sets.} Experimental results on the trilingual data sets WK3l are reported in this section. WK3l contains English (En), French (Fr), and German (De) knowledge graphs under DBpedia's ${\tt dbo\!\!:\!Person}$ domain, where a part of triples are aligned by verifying the ILLs on entities, and multilingual labels of the DBpedia ontology on some relations. The number of entities in each language is adjusted to obtain two data sets.
For each of the three languages thereof, WK3l-15k matches the number of nodes (about 15,000)
with FB15k---the largest monolingual graph used by many recent works~\cite{zhong2015aligning,lin2015learning,ji2015knowledge,jia2016locally}, and the number of nodes in 
WK3l-120k is several times larger.
For both data sets, German graphs are sparser than English and French graphs.
We also collect extra entity ILLs 
for the evaluation of cross-lingual entity matching, whose quantity is shown in Table~\ref{tbl:ills}.
Meanwhile, we derive another trilingual data set CN3l from ConceptNet~\cite{speer2013conceptnet}. Additional results on CN3l that lead to similar evaluation conclusions are reported in the Appendix of \cite{chen2016arxiv}.

\inv
\subsection{Cross-lingual Entity Matching} \label{sec:entity_matching}
The objective of this task is to match the same entities from different languages in $\kb$. Due to the large candidate \mbox{space}, this task emphasizes more on ranking a set of candidates rather than acquiring the best answer. We perform this task on both data sets to compare five variants of MTransE.\par

To show the superiority of MTransE, we adapt LM, CCA, and OT (which are introduced in Section~\ref{sec:related}) to their knowledge graph equivalences.

\stitle{Evaluation Protocol.} Each MTransE variant is trained on a complete data set. 
LM and CCA are implemented by inducing the corresponding transformations across separately trained knowledge models on monolingual graphs, while using the alignment sets as anchors.
Training OT is quite similar to MTransE, we add the process of orthogonalization to the training of the alignment model, since the regularization of vectors has already been enforced.
The entity ILLs are used as ground truth for test. We take these unidirectional links between English-French and English-German, i.e., four directions in total. For each ILL $(e, e')$, we perform a kNN search from the cross-lingual transition point of $e$ (i.e., $\tau(\mathbf{e})$)
and record the rank of $\mathbf{e'}$. Following the convention~\cite{xing2015normalized,jia2016locally},
we aggregate two metrics over all test cases, i.e., the proportion of ranks no larger than 10 $\hits$ (in percentage), and the mean rank $\mean$. We prefer higher $\hits$ and lower $\mean$ that indicate a better outcome. 
\par

For training,  we select the learning rate $\lambda$ among \{0.001, 0.01, 0.1\}, $\alpha$ among \{1, 2.5, 5, 7.5\}, $l_1$ or $l_2$ norm in loss functions, and dimensionality $k$ among \{50, 75, 100, 125\}. The best configuration on WK3l-15k is $\lambda=0.01$, $\alpha=5$, $k=75$, $l_1$ norm for Var$_1$, Var$_2$, LM, and CCA, $l_2$ norm for other variants and OT. While the best configuration on WK3l-120k is $\lambda=0.01$, $\alpha=5$, $k=100$, and $l_2$ norm for all models. The training on both data sets takes 400 epochs. \par

\stitle{Results.} We report $\hits$ and $\mean$ for WK3l-15k, and $\hits$ for WK3l-120k, on the four involved directions of cross-lingual matching in Table~\ref{tbl:matching}. As expected, without jointly adapting the monolingual vector spaces with the knowledge alignment, LM and CCA are largely outperformed by the rest. While the orthogonality constraint being too strong to be enforced in these cases, OT performs at most closely to the simplest cases of MTransE. For MTransE, Var$_4$ and Var$_5$ outperform the other three variants under all settings. The fairly close results obtained by these two variants indicate that the interference caused by learning an additional relation-dedicated transformation in Var$_5$ is negligible to the entity-dedicated transformation. Correspondingly, we believe that the reason for Var$_3$ to be outperformed by Var$_4$ and Var$_5$ is that it fails to differentiate well the over-frequent cross-lingual alignment from regular relations. Therefore, the characterization for cross-lingual alignment is negatively affected by the learning process for monolingual relations in a visible degree. 
Axis calibration appears to be unstable on this task. We hypothesize that this simple technique is affected by two factors: coherence between language-specific versions, and density of the graphs. Var$_2$ 
is always outperformed by Var$_1$ due to 
the negative effect of the calibration based on relations. We believe this is because multi-mapping relations are not so well-captured 
by TransE as 
explained in~\cite{wang2014knowledge}, therefore disturb the calibration of the entire embedding \mbox{spaces}. Although Var$_1$ still 
outperforms Var$_3$ on entity matching between English and French graphs in WK3l-15k, coherence somewhat drops alongside when scaling up to the larger data set so as to hinder the calibration. The German graphs are \mbox{sparse}, thus should have set a barrier for precisely constructing embedding vectors 
and hindered calibration on the other side. Therefore Var$_1$ still performs closely to Var$_3$ in the English-German task on WK3l-15k and English-French task on WK3l-120k, but is outperformed by Var$_3$ in the last setting. In general, the variants that use linear transformations are the most desired. 
This conclusion is supported
by their promising outcome on this task, and it is also reflected in the precision-recall curves shown in Figure~\ref{fig:matching}.

\begin{table*}[t!]
\begin{minipage}[t]{0.37\linewidth}
\centering
\captionsetup{justification=centering}
\caption{Accuracy of TWA verification (\%).}\label{tbl:alignment}
\vspace{-1em}
{\scriptsize
\begin{tabular}{c|cc|cc}
\bhline
Data Set & \multicolumn{2}{c|}{WK3l-15k} & \multicolumn{2}{c}{WK3l-120k}\\
\hline
Languages &  En\&Fr & En\&De & En\&Fr & En\&De\\
\bhline
LM&52.23&63.61&59.98&59.98\\
CCA&52.28&66.49&65.89&61.01\\
OT&93.20&87.97&88.65&85.24\\
\hline
Var${_1}$&93.25&91.24&91.27&91.35\\
Var${_2}$&90.24&86.59&89.36&86.29\\
Var${_3}$&90.38&84.24&87.99&87.04\\
Var${_4}$&94.58&\textbf{95.03}&\textbf{93.48}&93.06\\
Var${_5}$&\textbf{94.90}&94.95&92.63&\textbf{93.66}\\
\bhline
\end{tabular}
}
\end{minipage}
\hfill
\begin{minipage}[t]{0.3\linewidth}
\centering
\captionsetup{justification=centering}
\caption{Results of tail prediction ($\hits$).}\label{tbl:tail}
\vspace{-1em}
{\scriptsize
\begin{tabular}{c|cc|cc}
\bhline
Data Set & \multicolumn{2}{c|}{WK3l-15k} & \multicolumn{2}{c}{WK3l-120k}\\
\hline
Language & En & Fr & En & Fr\\
\bhline
TransE&\textbf{42.19}&25.06&36.78&25.38\\
Var${_1}$&40.37&23.45&39.09&25.52\\
Var${_2}$&40.80&24.77&36.02&21.13\\
Var${_3}$&40.97&22.26&35.99&19.69\\
Var${_4}$&41.03&25.46&\textbf{39.64}&\textbf{25.59}\\
Var${_5}$&41.79&\textbf{25.77}&38.35&24.68\\
\bhline
\end{tabular}
}
\end{minipage}
\hfill
\begin{minipage}[t]{0.3\linewidth}
\centering
\captionsetup{justification=centering}
\caption{Results of relation prediction ($\hits$).}\label{tbl:relation}
\vspace{-1em}
{\scriptsize
\begin{tabular}{c|cc|cc}
\bhline
Data Set & \multicolumn{2}{c|}{WK3l-15k} & \multicolumn{2}{c}{WK3l-120k}\\
\hline
Language & En & Fr & En & Fr\\
\bhline
TransE&61.79&62.55&60.06&65.29\\
Var${_1}$&60.18&60.73&\textbf{61.75}&65.46\\
Var${_2}$&54.33&62.98&61.11&61.47\\
Var${_3}$&58.32&59.44&60.14&48.06\\
Var${_4}$&63.74&\textbf{64.77}&60.26&\textbf{67.64}\\
Var${_5}$&\textbf{64.79}&63.71&60.77&66.86\\
\bhline
\end{tabular}
}
\end{minipage}
\vspace{-1.5em}
\end{table*}

\inv\inv
\subsection{Triple-wise Alignment Verification}

This task is to verify whether a given pair of aligned triples are truly cross-lingual counterparts. It produces 
a classifier that helps with verifying candidates of triple matching 
\cite{nguyen2011multi,rinser2013cross}. \par

\stitle{Evaluation Protocol.} We create positive cases by isolating 20\% of the alignment set. Similar to \cite{socher2013reasoning}, we randomly corrupt positive cases to generate negative cases. In detail, given a pair of correctly aligned triples $(T, T')$, it is corrupted by 
(i) randomly replacing one of the six elements in the two triples with another element from the same language, or (ii) randomly substituting either $T$ or $T'$ with another triple from the same language. 
Cases (i) and (ii) respectively contribute
negative cases that are as many as 100\% and 50\% 
of positive cases.
We use 10-fold cross-validation on these cases to train and evaluate the classifier.


We use a simple threshold-based classifier similar to the widely-used ones for triple classification \cite{socher2013reasoning,wang2014knowledge,lin2015learning}. For a given pair of aligned triples $(T, T')$ $=$ $\bigl((h, r, t), (h', r', t')\bigr)$, the dissimilarity function is defined as $f_d(T, T')$ $=$ $\left \| \tau (\mathbf{h})-\mathbf{h}'\right \|_2+\left \| \tau (\mathbf{r})-\mathbf{r}'\right \|_2+\left \| \tau (\mathbf{t})-\mathbf{t}'\right \|_2$. The classifier finds a threshold $\sigma$ such that $f_d<\sigma$ implies positive, otherwise negative. The value of $\sigma$ is determined by maximizing the accuracy for each fold on the training set. Such a simple classification rule adequately relies on how precisely each model represents cross-lingual transitions for both entities and relations.

We carry forward the corresponding configuration from the last experiment, just to show the performance of each variant under controlled variables.

\stitle{Results.} 
Table~\ref{tbl:alignment} shows the mean accuracy, 
with a \mbox{standard} deviation below 0.009 in cross-validation for all settings. Thus,
the results are statistically sufficient to reflect the performance of classifiers. Note that the results appear to be better than those of the previous task since this is a binary classification problem. Intuitively, the linear-transformation-based MTransE perform steadily and take the lead on all settings. We also observe that Var$_5$, though learns an additional relation-dedicated transformation, still performs considerably close to Var$_4$ (the difference is at most 0.85\%). The simple Var$_1$ is the runner-up, and is 
between 1.65\% and 3.79\% to the optimal solutions. However the relation-dedicated calibration in Var$_2$ causes a notable setback (4.12\%${\sim}$8.44\% from the optimal). The performance of Var$_3$ falls behind slightly more than Var$_2$ (4.52\%${\sim}$10.79\% from the optimal) due to the failure in distinguishing cross-lingual alignment from regular relations. Meanwhile, we single out the accuracy on the portion of negative cases where only the relation is corrupted for English-French in WK3l-15k. The five variants receive 97.73\%, 93.78\%, 82.34\%, 98.57\%, and 98.54\%, respectively. The close accuracy of Var$_4$ and Var$_5$ indicates that the only transformation learnt from entities in Var$_4$ is \mbox{enough} to substitute the relation-dedicated transformation in Var$_5$ for discriminating relation \mbox{alignment}, while learning the additional transformation in Var$_5$ does not notably interfere the original one. However, it 
applies differently to axis calibration since 
Var$_2$ does not improve but actually 
impairs the cross-lingual transitions for relations. For the same reasons as above, LM and CCA do not match with MTransE in this experiment as well, while OT performs closely to some variants of MTransE, but is still left behind by Var$_4$ and Var$_5$.
\par

\inv
\subsection{Monolingual Tasks}
The above experiments have shown the strong capability of MTransE in handling cross-lingual tasks. 
Now 
we report the results on comparing MTransE with its monolingual counterpart TransE on two monolingual tasks introduced in the literature~\cite{bordes2013translating,bordes2014open}, namely tail prediction (predicting $t$ given $h$ and $r$) and relation prediction (predicting $r$ given $h$ and $t$), using the English and French versions of our data sets.
Like previous works \cite{bordes2013translating,wang2014knowledge,jia2016locally},
for each language version, 10\% triples are selected 
as the test set, and 
the remaining 
becomes 
the training set. 
Each MTransE variant is trained upon both 
language versions of the training set for the knowledge model, while the intersection between the alignment set and the training set is used 
for the alignment models. TransE is 
trained on either language version of the training set. Again, we use the configuration 
from the previous experiment. 
\par

\stitle{Results.} The results for $\hits$ 
are reported 
in Tables~\ref{tbl:tail} and \ref{tbl:relation}.
They 
imply that MTransE preserves well 
the characterization of monolingual knowledge. 
For each setting, Var$_1$, Var$_4$, and Var$_5$ perform at least as well as 
TransE, and 
some even outperforms TransE under certain settings. 
This signifies that the alignment model does not interfere much with 
the knowledge model in characterizing monolingual relations,
but might have actually 
strengthened it since coherent portions of knowledge are unified by the alignment model.
\mbox{Since} such coherence is currently not measured, this question is left 
as a future work.
The other question that deserves further attention is,
how other 
knowledge models involving relation-specific entity transformations
~\cite{wang2014knowledge,lin2015learning,ji2015knowledge,jia2016locally,nguyenstranse} 
may influence monolingual and cross-lingual tasks.

\inv
\section{Conclusion and Future Work}
At the best of our knowledge, this paper is the first work that
generalizes 
knowledge graph embeddings to the multilingual scenario.
Our model MTransE characterizes monolingual relations and 
compares three different techniques to learn cross-lingual alignment for entities and \mbox{relations}.
Extensive experiments on the tasks of cross-lingual \mbox{entity} matching and triple alignment verification show that the linear-transformation-technique is the best 
among the three. 
Moreover,
MTransE preserves the key properties 
of monolingual knowledge graph embeddings on monolingual tasks
. \par

The results here are very encouraging, but we also point out opportunities for further work and improvements. In particular, we
should explore how to 
substitute the simple loss function of the knowledge model used in MTransE with more advanced 
ones involving relation-specific entity transformations.
More sophisticated 
tasks of cross-lingual triple completion may also be conducted.
Combining MTransE with multilingual word embeddings \cite{xing2015normalized}
is another meaningful direction since it will provide 
a useful tool to extract new relations from multilingual text corpora.

\bibliographystyle{named}
\begingroup
\bibliography{ref}
\endgroup
\clearpage


\newpage
\section{Appendix}
\subsection{Examples of Knowledge Alignment}		

We have already shown the effectiveness of MTransE in aligning cross-lingual knowledge, 		
especially the linear-transformation-based variants Var$_4$ and Var$_5$.
Now we discuss several examples 
to reveal insights on how our methods may be used in cross-lingual knowledge augmentation.		\par

\begin{table}[h]
\centering
\caption{Examples of cross-lingual entity matching.}
\label{table:case_study_entity_matching}
\vspace{-1em}		
\tiny			
\begin{tabular}{c|c|l}
\bhline
Entity       & Target   & Candidates (in ascending order of rank by Euclidean distance) \\
\bhline
\multirow{2}{*}{\pbox{10cm}{Barack\\ Obama}}
     & French   & \textbf{Barack Obama}, \emph{George Bush}, \emph{Jimmy Carter}, George Kalkoa\\
\cline{2-3}
     & German   & \textbf{Barack Obama}, \emph{Bill Clinton}, \emph{George h. w. Bush}, Hamid Karzai \\
\hline
\multirow{2}{*}{Paris}
     & French   & \textbf{Paris}, \emph{Amsterdam}, \textbf{\`a Paris}, \emph{Manchester}, De Smet\\
\cline{2-3}
     & German   & \textbf{Paris}, \emph{Languedoc}, \emph{Constantine}, \emph{Saint-maurice}, \emph{Nancy}\\
\hline
\multirow{2}{*}{California}
     & French   & \emph{San Francisco}, \emph{Los Angeles}, \emph{Santa Monica}, \textbf{Californie} \\
\cline{2-3}
     & German   & \textbf{Kalifornien}, \emph{Los Angeles}, \emph{Palm Springs}, \emph{Santa Monica}\\
\hline
\multirow{2}{*}{rock music}
     & French   & \emph{post-punk}, \textbf{rock alternatif}, \emph{smooth jazz}, \emph{soul jazz} \\
\cline{2-3}
     & German   & \textbf{rockmusik}, \emph{soul}, \emph{death metal}, \emph{dance-pop}\\
\bhline
\end{tabular}
\vspace{-1em}
\end{table}		
		
\begin{table}[h]
\centering
\vspace{-1em}
\caption{Examples of cross-lingual relation matching.}
\label{table:case_study_relation_matching}
\vspace{-1em}		
\tiny		
\begin{tabular}{c|c|l}
\bhline
Relation       & Target   & Candidates (in ascending order of rank by Euclidean distance)   \\
\bhline
\multirow{2}{*}{\pbox{10cm}{capital}}
     & French   & \textbf{capitale}, \emph{territoire}, pays accr\`editant, lieu de veneration \\
\cline{2-3}
     & German   & \textbf{hauptstadt}, \emph{hauptort}, \emph{gr\"undungsort}, \emph{city} \\
\hline
\multirow{2}{*}{\pbox{10cm}{nationality}}
     & French   & \textbf{nationali\'e}, \textbf{pays de naissance}, \emph{domicile}, \emph{r\'esidence}  \\
\cline{2-3}
     & German   & \textbf{nationalit\"at}, \textbf{nation}, letzter start, \emph{sterbeort} \\
\hline
\multirow{2}{*}{language}
     & French   & \textbf{langue}, r\'ealisations, lieu deces, \emph{nationalit\`e} \\
\cline{2-3}
     & German   & \textbf{sprache}, \textbf{originalsprache}, \textbf{lang}, \emph{land} \\
\hline
\multirow{2}{*}{nickname}
     & French   & \textbf{surnom}, 	descendant, texte, \emph{nom de ring} \\
\cline{2-3}
     & German   & \textbf{spitzname}, originaltitel, \emph{names}, \textbf{alternativnamen} \\
\bhline
\end{tabular}
\vspace{-1em}
\end{table}	

\begin{table}[h]
\centering
\vspace{-1em}
\caption{Examples of cross-lingual triple completion.}
\label{table:case_study_2}		
\vspace{-1em}		
\tiny			
\begin{tabular}{l|c|l}
\bhline
Query    & Target   & Candidates (in ascending order of rank) \\
\bhline
\multirow{2}{*}{\pbox{10cm}{(Adam Lambert, \\genre, ?\emph{t})}}
	& French  & \makecell[l]{\emph{musique ind\`ependante}, \textbf{musique alternative},\\ ode, \textbf{glam rock}} \\
\cline{2-3}
	& German  & \textbf{popmusik}, \textbf{dance-pop}, 	no wave, \emph{soul} \\
\hline
\multirow{2}{*}{\pbox{10cm}{(Ronaldinho, \\position, ?\emph{t})}}
	& French  & \textbf{milieu offensif}, \textbf{attaquant}, \emph{quarterback}, \emph{lat\`eral gauche}\\
\cline{2-3}
	&German& \textbf{st\"urmer}, \emph{linker fl\"ugel}, \textbf{angriffsspieler}, \emph{rechter fl¨¹gel}\\
\hline
\multirow{2}{*}{\pbox{10cm}{(Italy, ?\emph{r}, Rome)}}
	&French& \textbf{capitale}, \textbf{plus grande ville}, \textbf{chef-lieu}, garnison \\
\cline{2-3}
	&German& \textbf{hauptstadt}, \textbf{hauptort}, verwaltungssitz, stadion \\
\hline
\multirow{2}{*}{\pbox{10cm}{(Barack Obama, ?\emph{r},\\ George Bush)}}
    &French& \makecell[l]{\emph{ministre-pr\`esident}, \textbf{pr\`ed\`ecesseur}, \emph{premier ministre},\\ \emph{pr\`esident du conseil}} \\
\cline{2-3}
	&German& \textbf{vorg\"anger}, \textbf{vorg\"angerin}, besetzung, lied \\
\hline
\multirow{2}{*}{\pbox{10cm}{(?\emph{h}, instrument,\\ guitar)}}
    &French& \makecell[l]{\textbf{Brant Bjork}, \textbf{Chris Garneau}, \emph{David Draiman},\\ \textbf{Ian Mackaye}} \\
\cline{2-3}
	&German& \textbf{Phil Manzanera}, \emph{Styles P.}, \emph{Tina Charles}, \textbf{Luke Bryan} \\
\bhline
\end{tabular}
\vspace{-1em}
\end{table}
		
We start 
with the search of cross-lingual counterparts of entities and relations.		
We choose an entity (or relation) in English and then show the nearest candidates in French and German, respectively. 		
These candidates are listed by 
decreasing values 
of the Euclidean distance between their vectors in the target language space and the result point of cross-lingual transition.		
Several examples 
are shown in Table~\ref{table:case_study_entity_matching} and Table~\ref{table:case_study_relation_matching}.
In all tables of this subsection, we mark the exact answers as \textbf{boldfaced}, and the conceptually close ones as $italic$.
For example, in Table~\ref{table:case_study_entity_matching}, besides boldfacing the exactly correct answers for Barack Obama and Paris, we consider those who have also been U.S. presidents as conceptually close to Barack Obama, and European cities other than Paris as conceptually close to Paris. Also, in Table~\ref{table:case_study_relation_matching}, those French and German relations that have the meaning of settlements of significance are considered as conceptually close to \emph{capital}.
\par

We then move on to the more complicated cross-lingual triple completion task.		
We construct queries by replacing one element in an English triple with a question mark,
for which we seek for answers in another language.
Our methods need to transfer the remaining elements 
to the space of the target language		
and pick the best answer for the missing element.		
Table~\ref{table:case_study_2} shows some query answers.
It is noteworthy that the basic queries are already useful for aided cross-lingual augmentation of knowledge.
However, developing a joint model to support complex queries on multilingual knowledge graphs based on MTransE generated features appears to be a promising future work to support Q\&A on multilingual 
knowledge bases.
\par

\begin{figure*}[t]
\vspace{-1.5em}
\centering
\begin{subfigure}[c]{0.4\textwidth}
\centering
\includegraphics[width=\textwidth]{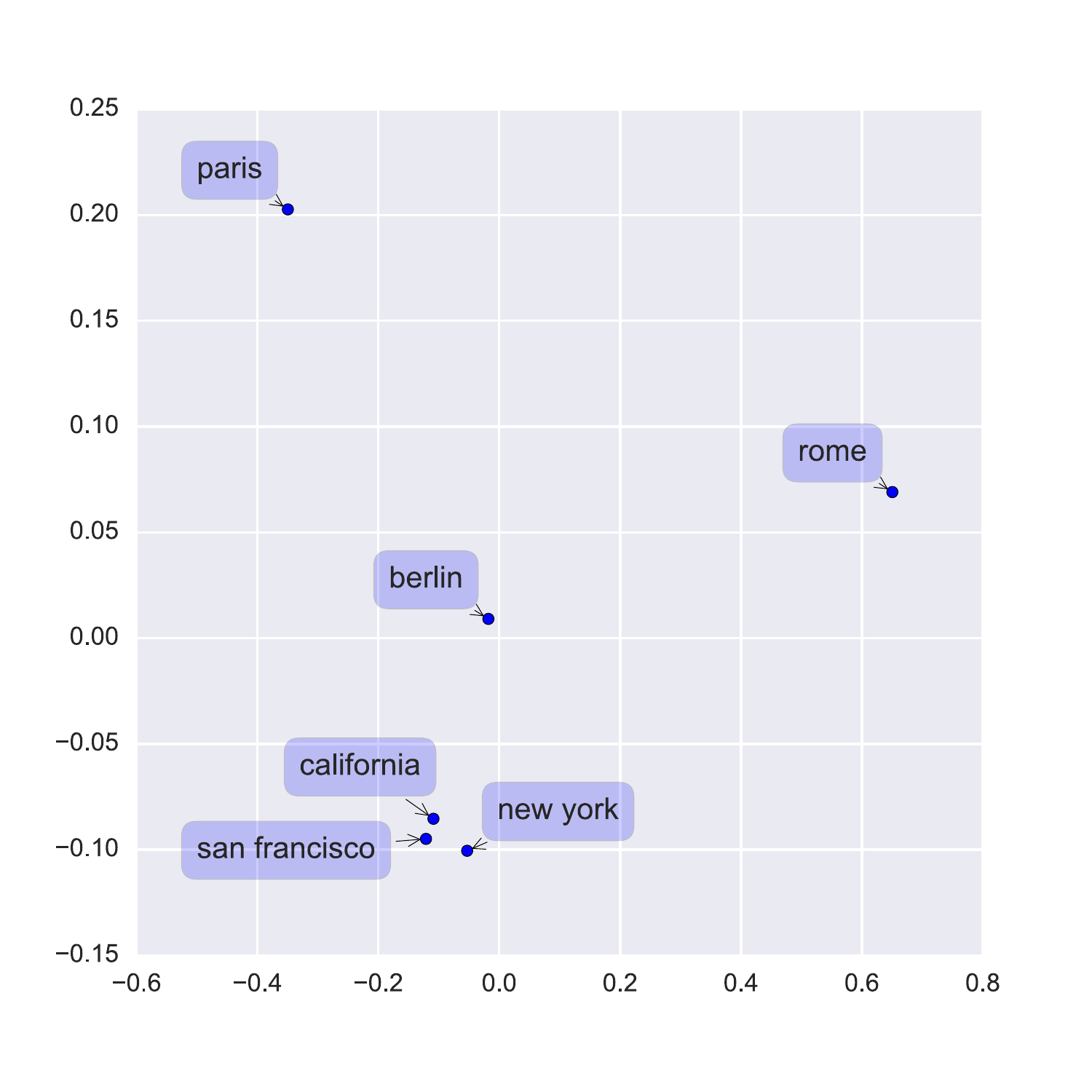}
\end{subfigure}
\hspace{5em}
\begin{subfigure}[c]{0.4\textwidth}
\centering
\includegraphics[width=\textwidth]{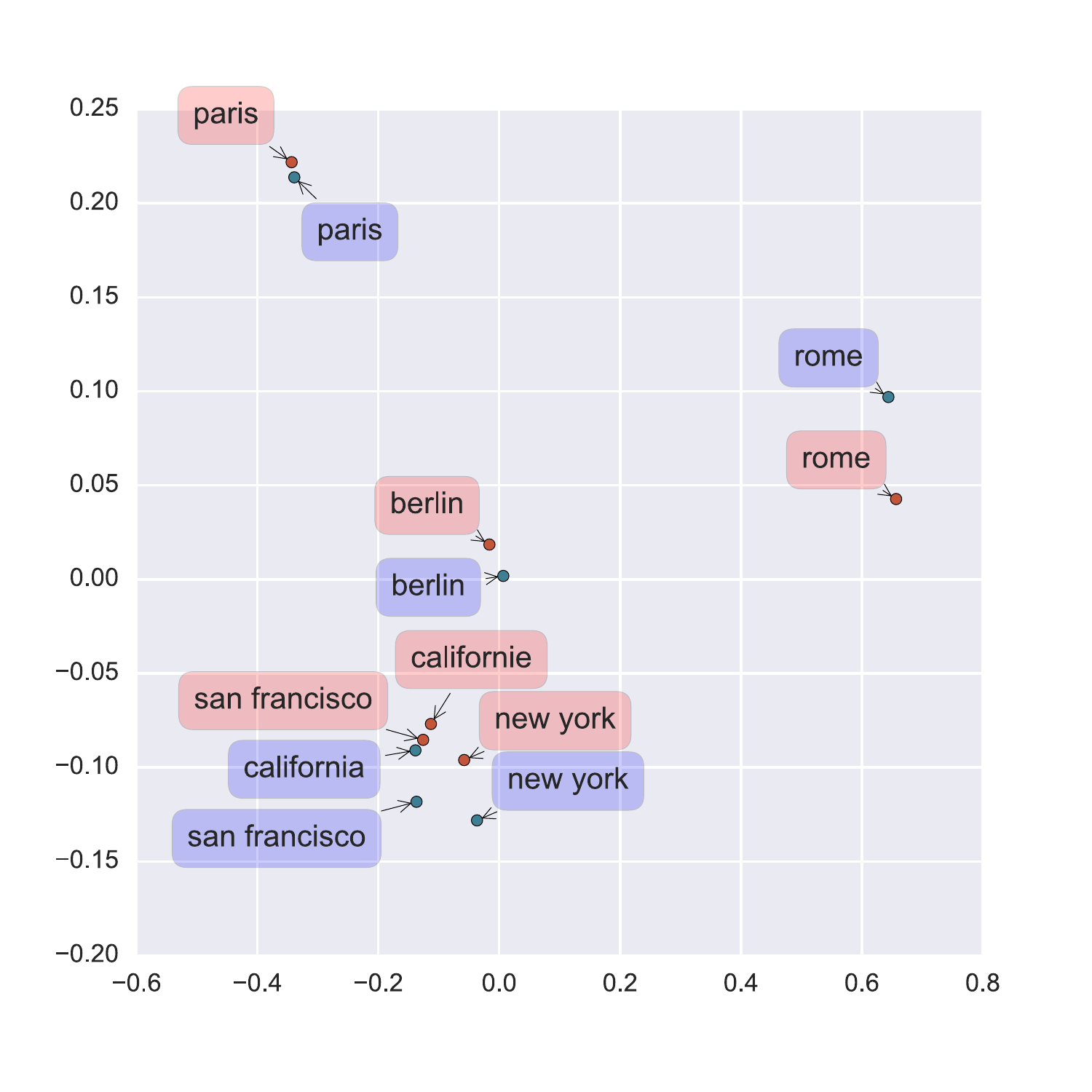}
\end{subfigure}
\vspace{-2em}
\caption{Visualization of
the result of Var$_4$ for
the same six English entities in their original space (left) and in French space after being transformed (right).
English entities are
rendered in blue, and
the corresponding French entities are
in light ruby.
}
\label{fig:viz}
\end{figure*}

Figure~\ref{fig:viz} shows the PCA projection of the {\em same} six English entities in their original English space and in French space after transformation.
We can observe that the vectors of English entities show certain structures, where the U.S. cities are grouped together and other countries' cities are well separated.
After transformation into French space, these English entities not only keep their original spatial emergence, but also are close to their corresponding entities in French.
This illustrates the transformation preserves mono-lingual structure and also it is able to capture cross-lingual information.
We believe this example illustrates the good performance we have demonstrated in cross-lingual tasks including cross-lingual entity matching and triple-wise alignment verification.
\par

\subsection{Additional Experimental Results}

\begin{table}[t!]
\centering
\caption{Statistics of the CN3l data set.}
\label{tbl:CN3l}
\vspace{-1em}
\scriptsize
\begin{tabular}{c|cccc}
\bhline
Type of triples&En triples&Fr triples&De triples&Aligned triples\\
\hline
Number of triples&47,696&18,624&25,560&\makecell{En-Fr:3,668\\En-De:8,588}\\
\bhline
Type of ILLs&En-Fr&Fr-En&En-De&De-En\\
\hline
Number of ILLs&2,154&2,146&3,485&3,813\\
\bhline
\end{tabular}
\vspace{-1em}
\end{table}


\begin{table*}[t!]
\centering
\begin{minipage}[t]{0.63\linewidth}
\captionsetup{justification=centering}
\caption{Cross-lingual entity matching (CN3l).}
\label{table:CN3l_entity_matching}
\scriptsize			
\begin{tabular}{c|cc|cc|cc|cc}
\bhline
Languages&\multicolumn{2}{c|}{En-Fr} & \multicolumn{2}{c|}{Fr-En} & \multicolumn{2}{c|}{En-De} & \multicolumn{2}{c}{De-En} \\
\hline
Metric&\multicolumn{1}{c}{Hits@10}&\multicolumn{1}{c|}{Mean}&\multicolumn{1}{c}{Hits@10}&\multicolumn{1}{c|}{Mean}&\multicolumn{1}{c}{Hits@10}&\multicolumn{1}{c|}{Mean}&\multicolumn{1}{c}{Hits@10}&\multicolumn{1}{c}{Mean}\\ \bhline
LM&\multicolumn{1}{c}{25.45}&\multicolumn{1}{c|}{1302.83}&\multicolumn{1}{c}{20.16}&\multicolumn{1}{c|}{1884.70}&\multicolumn{1}{c}{30.12}&\multicolumn{1}{c|}{954.71}&\multicolumn{1}{c}{18.04}&\multicolumn{1}{c}{1487.90}\\
CCA&\multicolumn{1}{c}{27.96}&\multicolumn{1}{c|}{1204.91}&\multicolumn{1}{c}{26.40}&\multicolumn{1}{c|}{1740.83}&\multicolumn{1}{c}{28.76}&\multicolumn{1}{c|}{1176.09}&\multicolumn{1}{c}{25.30}&\multicolumn{1}{c}{1834.21}\\
OT&\multicolumn{1}{c}{68.43}&\multicolumn{1}{c|}{42.30}&\multicolumn{1}{c}{67.06}&\multicolumn{1}{c|}{33.86}&\multicolumn{1}{c}{72.34}&\multicolumn{1}{c|}{74.98}&\multicolumn{1}{c}{69.47}&\multicolumn{1}{c}{44.38}\\
\hline
Var${_1}$&\multicolumn{1}{c}{61.37}&\multicolumn{1}{c|}{55.16}&\multicolumn{1}{c}{69.27}&\multicolumn{1}{c|}{33.60}&\multicolumn{1}{c}{63.06}&\multicolumn{1}{c|}{74.54}&\multicolumn{1}{c}{63.56}&\multicolumn{1}{c}{79.79}\\
Var${_2}$&\multicolumn{1}{c}{44.06}&\multicolumn{1}{c|}{226.63}&\multicolumn{1}{c}{57.15}&\multicolumn{1}{c|}{95.13}&\multicolumn{1}{c}{49.07}&\multicolumn{1}{c|}{219.97}&\multicolumn{1}{c}{49.15}&\multicolumn{1}{c}{214.58}\\
Var${_3}$&\multicolumn{1}{c}{73.73}&\multicolumn{1}{c|}{29.34}&\multicolumn{1}{c}{77.02}&\multicolumn{1}{c|}{14.82}&\multicolumn{1}{c}{70.55}&\multicolumn{1}{c|}{50.83}&\multicolumn{1}{c}{70.96}&\multicolumn{1}{c}{47.99}\\
Var${_4}$&\multicolumn{1}{c}{\textbf{86.83}}&\multicolumn{1}{c|}{\textbf{16.64}}&\multicolumn{1}{c}{\textbf{80.62}}&\multicolumn{1}{c|}{7.86}&\multicolumn{1}{c}{88.89}&\multicolumn{1}{c|}{\textbf{7.16}}&\multicolumn{1}{c}{\textbf{95.67}}&\multicolumn{1}{c}{\textbf{1.47}}\\
Var${_5}$&\multicolumn{1}{c}{86.21}&\multicolumn{1}{c|}{16.99}&\multicolumn{1}{c}{80.19}&\multicolumn{1}{c|}{\textbf{7.34}}&\multicolumn{1}{c}{\textbf{89.19}}&\multicolumn{1}{c|}{8.27}&\multicolumn{1}{c}{95.53}&\multicolumn{1}{c}{1.63}\\
\bhline
\end{tabular}
\end{minipage}
\hfill
\begin{minipage}[t]{0.3\linewidth}
\centering
\captionsetup{justification=centering}
\caption{Accuracy of triple-wise alignment verification (\%).}\label{tbl:alignmentcn}
\label{table:CN3l TWA}
{\scriptsize
\begin{tabular}{c|cc}
\bhline
Languages &  En\&Fr & En\&De\\
\bhline
LM&60.53&51.55\\
CCA&81.57&79.54\\
OT&93.01&87.59\\
\hline
Var${_1}$&93.92&91.89\\
Var${_2}$&87.30&82.70\\
Var${_3}$&88.95&84.80\\
Var${_4}$&\textbf{97.46}&\textbf{96.63}\\
Var${_5}$&97.18&95.42\\
\bhline
\end{tabular}
}
\end{minipage}
\vspace{-1em}
\end{table*}

We derive another data set CN3l from the MIT ConceptNet to evaluate MTransE, whose statistics are shown in Table~\ref{tbl:CN3l}.
Though being a smaller data set than WK3l-15k, knowledge graphs in CN3l are highly sparse.
Thereof, each language version of CN3l contains around 7,500 nodes and less than 41 types of relations.
The alignment sets are created based on the relation ${\tt TranslationOf}$ of the ConceptNet.
In this section we report the results of the two cross-lingual tasks on the CN3l data set for MTransE as well as all baselines.
Basically, these results lead to similar conclusions as we have on WK3l.

\stitle{Evaluation Protocol.} The metrics and evaluation procedures are the same as those on WK3l.
We select $\lambda$ among \{0.001, 0.01, 0.05\}, $\alpha$ among \{1, 2.5, 5, 7.5\}, $l_1$ or $l_2$ norm in loss functions, and dimensionality $k$ among \{25, 50, 75\}.
Optimal parameters are configured as $\lambda=0.001$, $\alpha=2.5$, $k=50$, and $l_1$ norm for all models.
To perform the evaluation under controlled variables, we again use one configuration on each model respectively in the two experiments.
Since the data set is smaller, training is limited to 200 epochs.

\stitle{Results of Cross-lingual Entity Matching.} The results are reported in Table~\ref{table:CN3l_entity_matching}. For the baselines, LM and CCA are again left far behind for being disjointedly trained. OT, however, takes the position ahead of Var$_1$. This is likely because the knowledge graphs in CN3l are highly sparse, therefore fewer interference of monolingual relations among entities makes the orthogonality constraint easier to fulfill. Even so, in all settings, OT is still largely outperformed by Var$_4$ and Var$_5$, which receives amazingly good results, thus steadily being the optimal solutions. Interestingly, Var$_3$ now ranks right behind the linear-transformation-based variants in most settings. This is quite reasonable because the cross-lingual transitions, which are regarded as a type of relation by Var$_3$ in the graphs, are now way less frequent in CN3l than they are in the much denser and more heterogeneous WK3l. Thus, this explains why it performs better than Var$_1$. For the same reason as we stated in Section~\ref{sec:entity_matching}, Var$_2$ is placed at last of the five MTransE variants in matching cross-lingual entities.

\stitle{Results of Triple-wise Alignment Verification.}  The results shown in Table~\ref{table:CN3l TWA} reflect the same conclusions of the experiment performed on WK3l that, the linear-transformation-based variants takes the lead ahead of the rest MTransE variants and the baselines. While Var$_1$, despite being the simplest, takes the second place with a satisfying accuracy in triple-wise alignment verification as well. The relation-dedicated calibration still causes interference in the optimization process of Var$_2$, therefore leads to a 4\%${\sim}$9\% drop of accuracy from Var$_1$. Var$_3$ performs slightly better than Var$_3$. On triple-wise alignment verification on CN3l, we receive exactly the same placement for evaluating the five MTransE variants. Meanwhile, for the baselines, OT is slightly worse than Var$_1$, CCA also receives acceptable accuracy which is however worse than all MTransE variants, while the accuracy of LM is just slightly better than random guessing.

\par
Above all, the results in CN3l indicates that MTransE also works promisingly on very sparse multilingual graphs, while the linear-transformation-based variants are the best representation techniques.

\end{document}